\documentclass{bmvc2k}

\usepackage{graphicx}
\usepackage{amsmath}
\usepackage{amssymb}
\usepackage{booktabs}
\usepackage{xcolor}
\usepackage{bm}
\usepackage{pifont}
\usepackage{enumitem}
\usepackage{soul}
\usepackage{dblfloatfix}
\usepackage{siunitx}

\usepackage{multirow}
\usepackage{xspace}
\usepackage{rotating}
\usepackage{framed}
\usepackage{bbm}
\usepackage{makecell}
\usepackage[export]{adjustbox}
\usepackage[capitalize]{cleveref}

\title{\scorename: A Novel Evaluation Score for \\Styled Handwritten Text Generation}

\addauthor{Vittorio Pippi}{vittorio.pippi@unimore.it}{}
\addauthor{Fabio Quattrini}{fabio.quattrini@unimore.it}{}
\addauthor{Silvia Cascianelli}{silvia.cascianelli@unimore.it}{}
\addauthor{Rita Cucchiara}{rita.cucchiara@unimore.it}{}

\addinstitution{
  University of Modena and Reggio Emilia\\
 Modena, IT
}

\newcommand{\scorename}{HWD}

\runninghead{Pippi et al.}{\scorename: A Novel Evaluation Score for Styled HTG}

\def\eg{\emph{e.g}\bmvaOneDot}

\def \ie {\emph{i.e.},}
\def \eg {\emph{e.g.},}

\newcommand{\tit}[1]{\smallbreak\noindent\textbf{#1.}}

\begin{document}

\maketitle

\begin{abstract}
Styled Handwritten Text Generation (Styled HTG) is an important task in document analysis, aiming to generate text images with the handwriting of given reference images. In recent years, there has been significant progress in the development of deep learning models for tackling this task. 
Being able to measure the performance of HTG models via a meaningful and representative criterion is key for fostering the development of this research topic. However, despite the current adoption of scores for natural image generation evaluation, assessing the quality of generated handwriting remains challenging. In light of this, we devise the Handwriting Distance (\scorename), tailored for HTG evaluation. In particular, it works in the feature space of a network specifically trained to extract handwriting style features from the variable-lenght input images and exploits a perceptual distance to compare the subtle geometric features of handwriting. Through extensive experimental evaluation on different word-level and line-level datasets of handwritten text images, we demonstrate the suitability of the proposed \scorename~as a score for Styled HTG. The pretrained model used as backbone will be released to ease the adoption of the score, aiming to provide a valuable tool for evaluating HTG models and thus contributing to advancing this important research area.
\end{abstract}

%-------------------------------------------------------------------------
\section{Introduction}
Styled Handwritten Text Generation (Styled HTG) entails producing realistic images of arbitrary handwritten text in a desired style given in the form of one or more exemplar style images. Those images can be used to: train models for document analysis tasks (\eg~Handwritten Text Recognition~\cite{zhang2019sequence, bhunia2019handwriting, kang2020distilling, cojocaru2021watch, bhunia2021metahtr, bhunia2021text, kang2021content, cascianelli2022boosting, pippi2023how}) in low-resource scenarios such as ancient languages or documents by specific authors; enhance the user experience in augmented reality scenarios and the public engagement at GLAM institutions (galleries, libraries, archives, and museums); assist physically impaired people in taking notes on electronic devices. 
Although HTG is receiving increased interest in recent years~\cite{fogel2020scrabblegan, kang2020ganwriting, bhunia2021handwriting, pippi2023handwritten}, a precise standard for evaluation has not yet been defined. 
It is important to note that evaluating the similarity between two writers' calligraphy involves more than considering the overall appearance of the text images. 
This concerns color and texture of background and ink, stroke thickness, slant, and roundness. Nonetheless, handwriting is characterized by the shape of individual characters, ligatures, and the spacing between characters in the text. Hence, a thorough evaluation procedure should consider all these factors to ensure an accurate and meaningful assessment of the ability of Styled HTG models to imitate a desired handwriting.

The commonly adopted approach is the one proposed by Kang et al.~\cite{kang2020ganwriting}, which entails exploiting the Fréchet Inception Distance (FID)~\cite{heusel2017gans}. The FID is computed on the features extracted by an Inception-v3 ConvNet trained on natural images from ImageNet. Thus, the FID is somehow unsatisfactory for measuring the faithfulness in the handwriting style of the generated images but rather captures the overall appearance~\cite{kang2020ganwriting, kang2021content, pippi2023handwritten}. Another critical point of applying the FID in the text images domain is that it leverages a backbone trained on images whose aspect ratio is very different from text images. These latter are usually wider than high, while natural images in ImageNet are roughly squared.  For this reason, in HTG evaluation, the FID is commonly computed on the beginning part of the text image, discarding the rest. This approach offers invariance with respect to the textual content but is prone to miss artifacts and dissimilarities that appear in the center or at the rightmost part of the image. Therefore, while the FID can help evaluate certain aspects of HTG, it does not provide a complete picture of the quality of the generated handwriting images. Moreover, different studies~\cite{binkowski2018demystifying, chong2020effectively} have shown that the FID is biased on the number of samples used for its calculation, resulting in lower values with more instances and higher and more unstable values when reducing the number of examples. 

To address these challenges, we propose a new evaluation score called Handwriting Distance (\scorename). The main characteristics of \scorename~consist of: 
(1) the use of robust style features extracted by a backbone trained on a large dataset of synthetic text images; 
(2) the application of a perceptual distance for style comparison; 
(3) the ability to handle variable-length text images; 
(4) the numerical stability even when computed on a limited number of samples. 
This is particularly useful in Styled HTG, where only a few real images per author are usually available. 
To assess the suitability of the proposed score for the Styled HTG task, we examine the values obtained when comparing sets of text images in the same style with respect to those obtained when comparing images in different styles. We demonstrate that the \scorename~effectively captures differences in the handwriting and is numerically stable. This makes it more suitable than the FID in expressing the performance of the Styled HTG models. Overall, the \scorename~could contribute to the field of HTG by providing a tailored and practical evaluation score to measure the realism and faithfulness of generated text images. 
The code and the weights of the convolutional backbone used to compute the \scorename~score can be found here: \url{https://github.com/aimagelab/HWD}.

\section{Related Work}
The early-proposed approaches to HTG ~\cite{wang2005combining, haines2016my} apply handcrafted geometric statistical-based feature extraction on human-made glyphs segmentations, then combine them with appropriate ligatures and render the results with texture and background blending. The two major limitations of these approaches are their inability to render glyphs and ligatures not observed for each style and their reliance on costly human intervention. In contrast, recent deep learning-based HTG approaches can infer styled glyphs even when they are not explicitly shown in the style examples. A majority of learning-based solutions are based on GANs~\cite{goodfellow2014generative}, either unconditioned (for Non-Styled HTG) or conditioned on a variety of handwriting style examples (Styled HTG). In the second scenario, style samples may consist of whole sentences or lines~\cite{davis2020text}, a few words~\cite{kang2020ganwriting,bhunia2021handwriting,pippi2023handwritten}, or a single word~\cite{gan2021higan,gan2022higan+,luo2022slogan}. 

The first learning-based Non-Styled HTG approach, proposed in~\cite{alonso2019adversarial}, entails generating fixed-sized images conditioned on the embedding of the desired textual content but does not control the calligraphic style of the output. Since, different from natural images, handwritten text images are highly variable-sized, subsequent approaches~\cite{fogel2020scrabblegan} entail concatenating character images. Styled HTG approaches condition the generation on both the text content and a vector representation of the style~\cite{kang2020ganwriting, davis2020text, gan2021higan, kang2021content, mattick2021smartpatch, gan2022higan+}. The two representations are obtained separately and then combined for generation, preventing those approaches from effectively capturing local writing style and patterns. On the other hand, the Transformer-based~\cite{vaswani2017attention} approach adopted in~\cite{bhunia2021handwriting, pippi2023handwritten} exploits the cross-attention mechanism between the style vector representation and the content text representation to entangle the content and the style, thus better rendering local style patterns. 

\tit{HTG Evaluation} As for the performance evaluation, models for HTG are evaluated by exploiting the FID~\cite{heusel2017gans}. The FID is a commonly used score for evaluating the quality of generative models. It exploits image representations extracted from an Inception-v3, which are fit to two multivariate Gaussian distributions, one from real images and the other from generated images. For this reason, the FID tends to focus more on general image characteristics rather than the shape of handwriting. Furthermore, the backbone network used to compute the FID is trained on ImageNet, which contains natural images whose domain and aspect ratio are completely different from those of handwritten text images, which can result in misleading values. Other adopted metrics are the Geometric Score~\cite{khrulkov2018geometry} and the Character Error Rate. The latter measures the readability of the generated text images, which serves as a proxy to express their realism, \ie~how similar to a well-formed text they look. However, as discussed in~\cite{kang2020ganwriting,kang2021content,pippi2023handwritten}, these measures fail to capture all the desired characteristics of a well-generated styled text image. In this work, we propose a score specific for evaluating HTG models to address this point.
\section{Handwriting Distance}
\begin{figure}
    \centering
    \includegraphics[width=\textwidth]{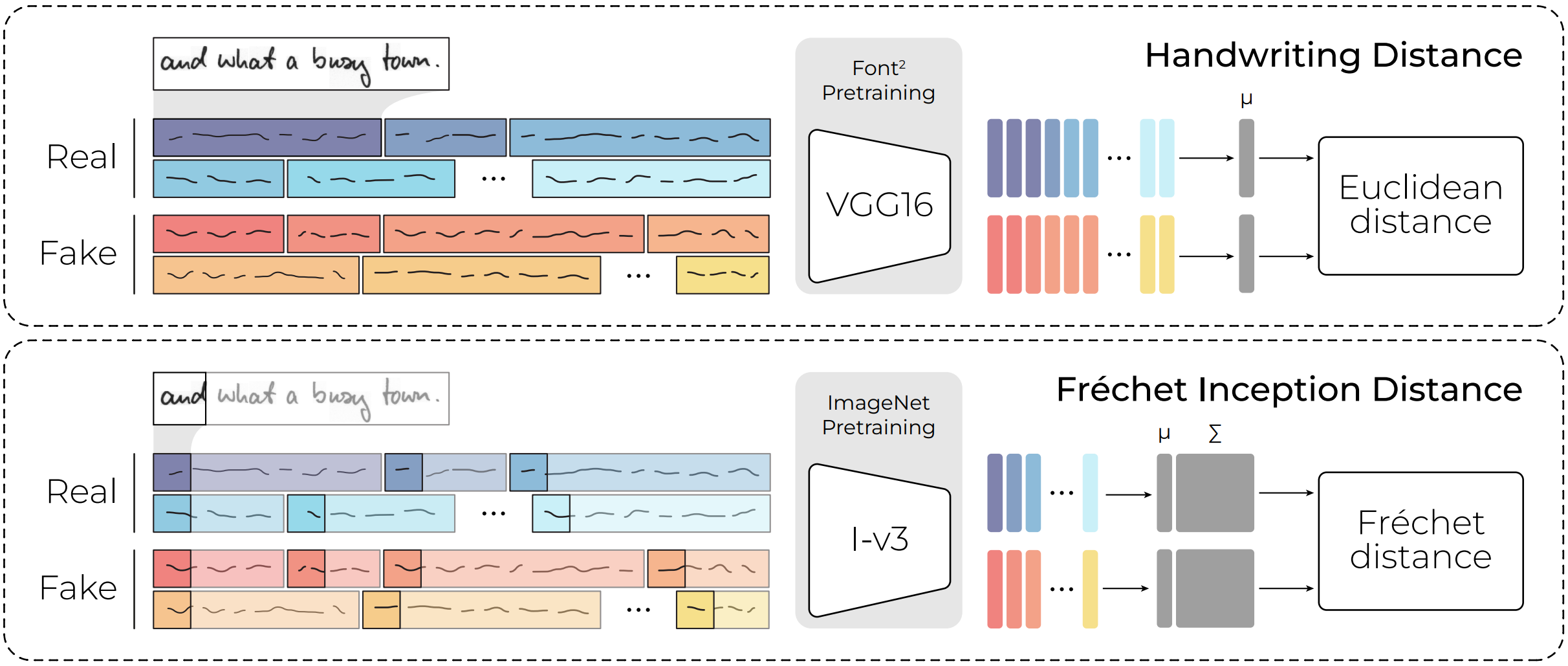}
    \caption{Pipelines for the computation of the proposed~\scorename~score (top) and the commonly used FID score (bottom) from a set of real and generated text images with varying widths. Top: the samples are processed by a VGG16 pretrained on a synthetic dataset of text images. The different widths of the inputs result in a different number of feature vectors. The value of \scorename~is the Euclidean distance between the feature vectors' means $\mu$. Bottom: the text images are cropped to a square and fed to an Inception-v3 pretrained on ImageNet, obtaining one feature vector per image. The FID is the Fréchet distance between the distributions of the features, represented as multivariate Gaussians with mean $\mu$ and covariance matrix $\Sigma$.
    }\vspace{-0.5cm}
    \label{fig:pipeline}
\end{figure}

Inspired by the strategy adopted for natural image generation evaluation, Styled HTG works employ the FID score, adapted as described in~\cite{kang2020ganwriting} and depicted in Figure~\ref{fig:pipeline} (bottom). In this work, we devise an alternative score for evaluating the performance of Styled HTG models, called \scorename. The main characteristics of \scorename~are the domain-aware image representation strategy and the use of a perceptual distance instead of a distribution distance. The pipeline for computing our proposed score is described below and depicted in Figure~\ref{fig:pipeline} (top).

\subsection{Text Images Representation}
When evaluating Styled HTG, we consider the set of real images $\mathbf{X}_{m}{=}\{\mathbf{x}_{m, i}\}_{i=0}^N$, where $N$ is the number of samples for writer $m$, and the set $\mathbf{X'}_{m}{=}\{\mathbf{x'}_{m, i}\}_{i=0}^{N'}$ of generated images in the style of writer $m$. In general, the number of generated images, $N'$, can differ from $N$. 

\tit{Domain-Specific Feature Extraction} 
Given the constrained domain of the images for the HTG task, we propose to use a domain-specific backbone as feature extractor. In particular, we adopt a VGG16 pretrained on Font$^2$, a large synthetic dataset of text images built according to the procedure presented in~\cite{pippi2023evaluating}. We choose VGG16 as a backbone for its superiority in extracting meaningful style representations compared to deeper networks featuring skip connections~\cite{wang2021rethinking}. The pretraining dataset contains more than 100M samples obtained by rendering 10400 English words in 10400 calligraphic fonts and superimposing them to paper-like backgrounds. Random geometric distortions and color transformations are also performed to increase the realism of the images. We train the VGG16 backbone to classify the images according to their calligraphic font. The high visual variability of the datasets forces the network to learn features that represent the handwriting given by the font and disregard the overall visual appearance. In this way, the features extracted by our adopted backbone are strong representations of the handwriting style. For computing the \scorename, we feed the pretrained VGG16 with the whole image resized to a height of 32 pixels while keeping the aspect ratio. Note that since the images have different widths, we feed them to the network one by one to avoid using padding. 

\tit{Variable-lenght Images Representation} 
To handle the difference in the aspect ratio of the natural images in ImageNet and text images, the strategy adopted in~\cite{kang2020ganwriting} and the following HTG approaches entails feeding to the Inception-v3 backbone only a squared, truncated sample. The text image is reshaped to a height of 32 pixels, keeping the aspect ratio, and then cropped at the beginning to have a fixed width (32 pixels). This strategy has two drawbacks. First, it disregards a large part of the text image. Second, the characters appearing more often at the beginning of words due to language regularity assume more importance than others in computing the score. Consequently, this strategy prevents evaluating the ability of the HTG models to generate images of variable-length texts with the same quality throughout the entire image. To overcome these limitations, we propose processing the whole images with the VGG16 backbone for the \scorename. The representations are obtained from the feature maps of the last convolutional block.  
Such feature maps have shape $1\times W\times 512$, where $W$ depends on the input text image width. From those maps, we obtain a set of $512$-sized feature vectors that represent the images. As a result, wide images are represented by a larger set of vectors than short ones. When computing the HWD for writer $m$, all the vectors from all the real images sets are gathered and then averaged (the same applies to the generated images). In the gathering set, there will be more vectors coming from the longer words, and thus, the longer words will have a bigger impact on the mean of the gathering set. 
In summary, each image for writer $m$, both real and generated, is represented by a set of vectors. The real images are represented by sets as $\mathbf{y}_{m, i}{=}\{f(\mathbf{x}_{m, i})_j\}_{j=0}^{W_i}$, and the generated ones by $\mathbf{y'}_{m, i}{=}\{f(\mathbf{x'}_{m, i})_j\}_{j=0}^{W_i'}$, where $f(\cdot)$ denotes the feature extraction, and $W_i$ and $W_i'$ are the number of vectors extracted from the $i$-th real and generated images, respectively. 

\subsection{Perception-Aware Feature Distance}
The main idea behind the distribution distance-based evaluation scores, such as the FID, is to evaluate the performance of a generative model by its ability to generate images that match the distribution of the real ones. Our domain of interest is more constrained than the generation of natural images. In fact, Styled HTG entails considering the handwriting, expressed by subtle geometric features other than macroscopic texture. In light of this, we argue that a score capturing the perceptual aspects is more suitable than one based on the distance between feature distributions. Therefore, we employ the Euclidean distance between the averaged feature vectors of the real and generated images in the style of the same writer:
\begin{equation*}
    Y_m = \frac{\sum_{i=1}^{N} \sum_{j=1}^{W_i} f(\mathbf{x}_{m, i})_j}{\sum_{i=1}^{N}W_i}
    \qquad
    \text{and}
    \qquad
    Y'_m = \frac{\sum_{i=1}^{N'} \sum_{j=1}^{W'_i} f(\mathbf{x'}_{m, i})_j}
   {\sum_{i=1}^{N'}W'_i}.
\end{equation*}
For the images in the style of writer $m$, the \scorename~is given by:
\begin{equation*}
    \text{\scorename}_m = \lVert Y_m - Y'_m \rVert_2.
\end{equation*}
Note that when computed on robust image representations, \eg~obtained from a backbone trained on a semantic prediction task, the Euclidean distance is highly predictive of the perceptual similarity between images~\cite{zhang2018unreasonable}.
Finally, the \scorename~on datasets containing images in the style of $M$ different authors is obtained as
\begin{equation*}
    \text{HWD} = \frac{1}{M}\sum_{m=i}^{M} \text{\scorename}_m.
\end{equation*}
The \scorename~score has the non-negativity, symmetry, triangular inequality properties, but it is not guaranteed that it exhibits the identity of indiscernible elements property since two non-identical images might have $\text{HWD}=0$ due to their representation via the pre-trained backbone. This is a desirable characteristic for the HTG task, in which text images containing different texts and different backgrounds should be at low (or even zero) HWD if they are written in the same handwriting style.
\section{Experimental Analysis}
For our experimental analysis of the proposed \scorename~score, we consider images from a number of multi-author and single-author datasets. We compare the \scorename~score against the FID score in the variant proposed in~\cite{kang2020ganwriting}, which is the common approach adopted in Styled HTG. The comparison is performed along two dimensions. First, we assess its capability to recognize corresponding handwriting styles, quantifying the style verification capability with the Overlap coefficient and the Equal Error Rate (EER). Second, we compare the numerical stability of the proposed \scorename~score and the FID. Additionally, we evaluate current Styled HTG State-of-the-Art models with our score and other common metrics for image generation evaluation. Finally, we conduct extensive ablation analyses on the main components of our approach to investigate their individual contributions. Further results can be found in the Supplementary material.

\subsection{Considered Datasets}\label{ssec:mw_dataset}
The considered multi-author datasets are described below and in Table~\ref{tab:scores_on_datasets}.
\tit{IAM} The IAM Database~\cite{marti2002iam} is a collection of greyscale document scans by 657 writers. These are written in English with ink on white paper and cleaned digitally. Here, we use the line-level version of the dataset.
\tit{RIMES} The RIMES Database~\cite{augustin2006rimes} consists of binary French documents by 1500 authors. For this dataset, we use the line-level version.
\tit{CVL} The CVL Database~\cite{kleber2013cvl} features word images obtained from RGB scans of English and German manuscripts, written with ink on white paper by 310 writers.
\tit{KHATT} The KHATT Database~\cite{mahmoud2014khatt} contains binarized images of handwritten Arabic words handwritten by 838 people.
\tit{BanglaWriting} The BanglaWriting Dataset~\cite{mridha2021banglawriting} is composed of greyscale images of Bengalese words handwritten by 212 authors. 
\tit{NorHand} The NorHand Dataset~\cite{maarand2022comprehensive} features text lines extracted from greyscale scans of ancient documents written with ink on yellowed paper by 12 Norwegian authors.
\\ 
The considered single-author datasets are presented below.
\tit{Saint Gall} The Saint Gall Dataset~\cite{fischer2011transcription} features binary images of 1410 lines from a medieval manuscript written in Latin with gothic calligraphy.
\tit{Washington} The George Washington Dataset~\cite{fischer2012lexicon} contains binary images of 656 lines from English letters written by American President George Washington.
\tit{Rodrigo} The Rodrigo Database~\cite{serrano2010rodrigo} contains 20357 lines extracted from greyscale scans of an historical manuscript, written in Spanish with ink on ancient paper.
\tit{ICFHR14} The ICFHR14 Dataset~\cite{sanchez2014icfhr2014} is a collection of 11473 lines extracted from greyscale scans of ancient pages, written by the English philosopher Jeremy Bentham.
\tit{Leopardi} The Leopardi Dataset~\cite{cascianelli2021learning} is a collection of 2459 lines from RGB scans of letters by the Italian poet Giacomo Leopardi, written with ink on ancient paper.
\tit{LAM} The LAM Dataset~\cite{cascianelli2022lam} includes 25823 text lines images obtained from RGB scans of ancient letters in Italian, written by Lodovico Antonio Muratori.

\subsection{Sensitivity to the Handwriting}
We evaluate the sensitivity of \scorename~to the handwriting style by splitting the multi-author datasets. We consider half of the samples for each featured writer as references and the other half as if they were the output of a perfect Styled HTG model. Then, we compare the distributions of the \scorename~and the FID computed on text images of multiple matching and non-matching authors pairs. Note that, in such an ideal case for Styled HTG, both the HWD and the FID should be as close as possible to their best value. Therefore, the more these two distributions are separated, the better the corresponding score captures the handwriting similarity between the considered images. 

\begin{figure}
    \centering
    \includegraphics[width=\textwidth]{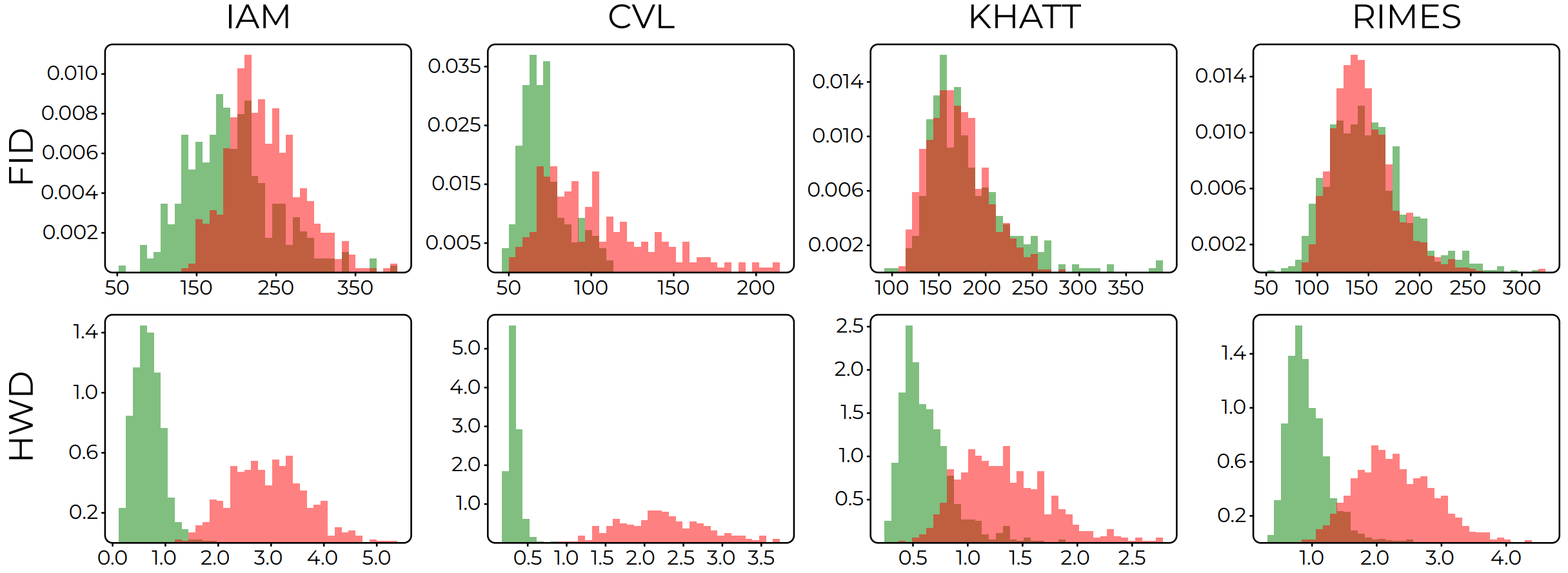}
    \caption{Distributions of different scores used to evaluate HTG models when applied on same-author (green) or different-author (red) subsets. The overlap area is in dark red.
   }
    \label{fig:distributions}
\end{figure}

The obtained distributions are reported in Figure~\ref{fig:distributions}. 
We observe that the histograms for the FID show significant overlap and that there is no clear separation between the distributions of matching and non-matching authors pairs. Moreover, except for CVL, which has a higher number of samples per author compared to the other three datasets, the FID values of the corresponding authors’ distribution are roughly above 100. Such high FID values highlight the bias of the score when computed on a few samples. On the other hand, the histograms for the \scorename~are more separated. 

In addition, we quantitatively evaluate the style recognition capability by comparing the distributions in Figure~\ref{fig:distributions} in terms of the Overlap coefficient, a statistical index that quantifies the overlap between distributions, and the EER, which represents the point where the False Acceptance Rate equals the False Rejection Rate. 
The results are reported in Table~\ref{tab:scores_on_datasets}. 
As we can see, for both the FID and HWD, the Overlap and ERR are lower on the IAM and CVL datasets than in KHATT and RIMES. We argue the causes to be, respectively, the languages of the datasets and the similar overall appearance of the samples from the same author. In fact, the FID mainly focuses on the latter, while HWD performs feature extraction with a VGG16 pretrained on Font$^2$, which contains words in the same language as IAM and CVL. Nonetheless, the HWD score achieves lower Overlap and EER values than the FID in all datasets, including those in languages different from English.

\begin{table}[]
    \centering
    \resizebox{\linewidth}{!}{
    \setlength{\tabcolsep}{.35em}
    \begin{tabular}{lc c crr rr c rr}
    \toprule
        & &
        \multirow{2}{*}{\textbf{Language}} & 
        \multirow{2}{*}{\textbf{Samples}} & 
        \multirow{2}{*}{\textbf{Authors}} & 
        \multirow{2}{*}{\makecell{\textbf{Avg. Samples}\\ \textbf{per author}}} & 
        \multicolumn{2}{c}{\textbf{FID}} &&
        \multicolumn{2}{c}{\textbf{\scorename}}\\[-2.5pt]
        \cmidrule(lr){7-8} \cmidrule(lr){10-11}\\[-13.5pt]
        &&&&&& \textbf{Overlap} & \textbf{EER} && \textbf{Overlap} & \textbf{EER} \\ 
    \midrule
        \textbf{Norhand}       && Norwegian      & 21939  &   12~~~ & 1828.25~~~~ &  4.2~~~ &  4.2~ &&  0.0~~~ & 0.0~ \\
        \textbf{BanglaWriting} && Bengali        & 17265  &  212~~~ &   81.44~~~~ & 11.6~~~ &  5.6~ &&  6.1~~~ & 2.9~ \\
        \textbf{CVL}           && English/German & 13473  &  310~~~ &   43.46~~~~ & 24.7~~~ & 12.5~ &&  0.0~~~ & 0.0~ \\
        \textbf{IAM}           && English        & 13353  &  657~~~ &   20.32~~~~ & 27.1~~~ & 13.6~ &&  0.7~~~ & 0.3~ \\
        \textbf{KHATT}         && Arabic         & 11427  &  838~~~ &   13.64~~~~ & 40.3~~~ & 21.6~ && 12.0~~~ & 5.9~ \\
        \textbf{RIMES}         && French         & 12111  & 1500~~~ &    8.07~~~~ & 39.1~~~ & 20.8~ &&  7.0~~~ & 3.3~ \\
    \bottomrule
    \end{tabular}}
    \caption{Overlap and EER of the FID and \scorename~values calculated on the images in the considered multi-author datasets.}
    \label{tab:scores_on_datasets}
\end{table}

\subsection{Sensitivity to the Number of Samples}
\begin{figure}
    \centering
    \includegraphics[width=\linewidth]{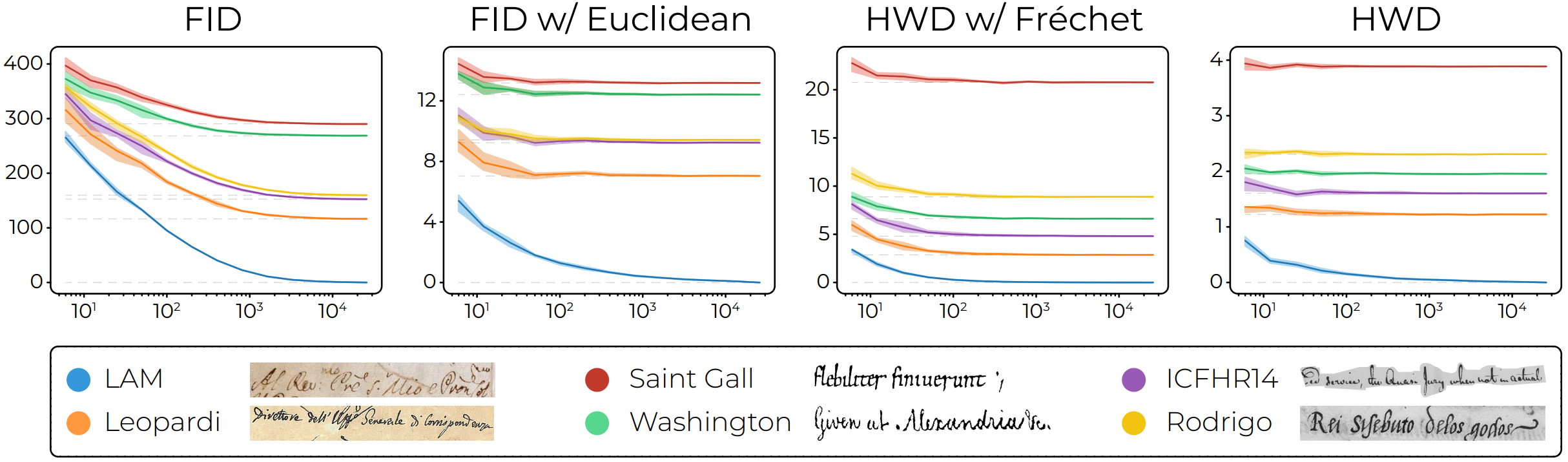}
    \caption{Left to right: comparison between FID, a version of the FID exploiting the Euclidean distance (FID w/ Euclidean), a version of \scorename~exploiting the Fréchet Distance (\scorename~w/ Fréchet), and \scorename~with varying number of samples. The lines denote the mean, and the transparent bands represent the range between the 25th and 75th percentiles, obtained with 10 calculation runs.}
    \label{fig:sample_size}
\end{figure}

Numerical stability is an important factor to consider when assessing a score. As argued by~\cite{chong2020effectively,binkowski2018demystifying}, the FID exhibits a strong bias towards the number of samples. To assess the stability of \scorename, we consider the large single-author LAM dataset and compute the values of the \scorename~and FID on images from LAM against variably sized subsets of images from ICFHR14, Saint Gall, Leopardi, Rodrigo, Washington, and LAM itself. We determine the mean of the scores over multiple runs and also consider the range between the 25th and 75th percentiles of values. 
The results are reported in Figure~\ref{fig:sample_size}. 

It can be observed that the FID (left-most plot) shows a significant bias towards the number of images: all the curves start with very high values and decrease slowly until at least around 2000 samples are used to obtain the score. On the other hand, the \scorename~(right-most plot) is more stable with respect to the subset sizes, reaching a plateau even when computed on around 100 samples. 
To further investigate the cause of this behavior, we compute the Euclidean distance on the Inception-v3 features (FID w/ Euclidean) and the Fréchet distance on the VGG16 features (HWD w/ Fréchet). In both cases (center-left plot and center-right plot of Figure~\ref{fig:sample_size}, respectively), we can observe a slight bias with respect to the number of samples used for the computation. 
Moreover, the computation of these scores necessitates roughly one more order of magnitude of images to reach a plateau, compared to the case of the FID and the \scorename, respectively. For the FID w/ Euclidean, we argue that this is because its Inception-v3 backbone is applied to the beginning part of the text images. As a result, more samples are needed to effectively represent the author’s handwriting. In the case of the HWD w/ Fréchet, we argue that this score suffers from the numerical instability of the Fréchet Distance, which fits the image representations to multivariate Gaussians.

Further, by comparing the values in the plots in the sight of the exemplar images reported in Figure~\ref{fig:sample_size}, we can make some considerations on the convolutional backbones used as feature extractors.
Inception-v3-based scores assign dataset distances according to their overall appearance. For instance, ICFHR14 and Rodrigo (both containing greyscale images) are close to each other, and the same happens for Saint Gall and Washington (both containing binarized images). On the other hand, the VGG16-based scores focus more on the handwriting. Thus, Saint Gall is isolated from the others because of its peculiar gothic calligraphy, while ICFHR14 is closer to Washington and Leopardi than to Rodrigo, reflecting the similarity in the cursive calligraphies. 
This characteristic is better suited for the Styled HTG task, as we are interested in evaluating the ability of a model to mimic the handwriting style.

\subsection{Styled HTG Models Evaluation}
For reference, we report the performance of State-of-the-Art Styled HTG models trained on the IAM dataset. In particular, we consider the conditional-GAN-based HIGAN+ approach~\cite{gan2022higan+}, which uses a single image to condition the generation and disentangles text and style,  and the few-shot Convolutional-Transformer-based models HWT~\cite{bhunia2021handwriting} and  VATr~\cite{pippi2023handwritten}, which capture both global and character-level style-text dependencies by exploiting cross-attention. The latter approach represents the textual input with visual archetypes (\ie~Unifont-rendered characters) instead of one-hot vectors, as done in the other two. 
In Table~\ref{tab:scores_on_words}, we evaluate the aforementioned models in terms of the proposed HWD and other scores for image generation evaluation. In particular, we consider the FID score; the Geometric Score (GS)~\cite{khrulkov2018geometry}, which compares the data and model manifold estimates; and the Kernel Inception Distance (KID)~\cite{binkowski2018demystifying}, which relaxes the Gaussian assumption of the FID and uses the Maximum Mean Discrepancy to compare the distributions of the features.

\begin{table}[]
    \centering
    \resizebox{.55\linewidth}{!}{
    \begin{tabular}{lc cccc}
    \toprule
        & 
        \textbf{FID} & 
        \textbf{KID} & 
        \textbf{GS} & 
        \textbf{\scorename} \\ 
    \midrule
        \textbf{HWT}  &  23.36 & 1.37$\times10^{-2}$ & \textbf{1.05$\mathbf{\times10^{-2}}$} & 1.928 \\
        \textbf{HIGAN+} &  \textbf{18.21} & 9.38$\times10^{-3}$ & 2.15$\times10^{-2}$ & 1.237 \\
        \textbf{VATr}    &  18.80 & \textbf{7.06$\mathbf{\times10^{-3}}$} & 2.19$\times10^{-2}$ & \textbf{0.828} \\
    \bottomrule
    \end{tabular}}
    \caption{Scores of the considered HTG models when generating the same or different words as those in the style reference images of the IAM test set. Best performance in bold.}
    \label{tab:scores_on_words}
\end{table}

\subsection{Ablation Analysis}
Finally, we analyze the effects of the four main components of our proposed score: the backbone, the pretraining dataset, the input image portion, and the distance measure. We perform the ablation analysis on the IAM dataset and report the results in Table~\ref{tab:ablation}. Looking at the results, we notice that the backbone used to extract the features plays a crucial role in the separability scores, as the use of VGG16 leads to very good results also when pretrained on ImageNet. A second important aspect is the feature distance metric used. The Fréchet distance on the VGG16-extracted features achieves good results on all settings. Nevertheless, it is not influenced by the backbone pretraining and the image portion used to extract the features. On the other hand, the Euclidean distance fully exploits the input information (\ie~the quantity and the type of feature vectors) and thus is the best in the \scorename~setting.  

\begin{table}[]
    \centering
    \resizebox{0.7\linewidth}{!}{
    \begin{tabular}{cccc rr}
        \toprule
        \multirow{2}{*}{\textbf{Backbone}} & 
        \multirow{2}{*}{\makecell{\textbf{Pretraining}\\ \textbf{Dataset}}} & 
        \multirow{2}{*}{\makecell{\textbf{Image}\\ \textbf{Portion}}} & 
        \multirow{2}{*}{\textbf{Distance}} &
        \multicolumn{2}{c}{\textbf{IAM}} \\[-2.5pt]
        \cmidrule(lr){5-6} \\[-13.5pt]
         & & & & \textbf{Overlap} & \textbf{EER} \\ 
        \midrule
        Inception-v3 & ImageNet & Beginning & Fréchet   & 27.1~~~ & 13.6~ \\
        Inception-v3 & ImageNet & Beginning & Euclidean & 29.6~~~ & 14.5~ \\
        Inception-v3 & ImageNet & Whole & Fréchet       & 24.0~~~ & 11.6~ \\
        Inception-v3 & ImageNet & Whole & Euclidean     &  8.5~~~ &  3.9~ \\
        \midrule
        Inception-v3 & Font$^2$ & Beginning & Fréchet   & 18.8~~~ &  9.3~ \\
        Inception-v3 & Font$^2$ & Beginning & Euclidean & 11.3~~~ &  4.8~ \\
        Inception-v3 & Font$^2$ & Whole     & Fréchet   & 19.0~~~ &  9.1~ \\
        Inception-v3 & Font$^2$ & Whole     & Euclidean &  7.2~~~ &  3.3~ \\
        \midrule
        VGG16        & ImageNet & Beginning & Fréchet   &  3.2~~~ &  1.6~ \\
        VGG16        & ImageNet & Beginning & Euclidean & 26.2~~~ & 13.0~ \\
        VGG16        & ImageNet & Whole     & Fréchet   &  2.8~~~ &  1.2~ \\
        VGG16        & ImageNet & Whole     & Euclidean &  6.2~~~ &  2.9~ \\
        \midrule
        VGG16        & Font$^2$ & Beginning & Fréchet   &  3.4~~~ & 1.7~ \\
        VGG16        & Font$^2$ & Beginning & Euclidean & 16.5~~~ & 8.2~ \\
        VGG16        & Font$^2$ & Whole     & Fréchet   &  3.5~~~ & 1.6~ \\
        VGG16        & Font$^2$ & Whole     & Euclidean & \textbf{0.7}~~~ & \textbf{0.3}~ \\
        \bottomrule
    \end{tabular}}
    \caption{Ablation analysis of the main components of the \scorename~score. Note that the first row is the FID while the last is the complete \scorename. Best performance in bold.}\label{tab:ablation}
\end{table}

\subsection{Sensitivity to the Visual Appearance}
\begin{figure}[h]
    \centering
    \includegraphics[width=.8\linewidth]{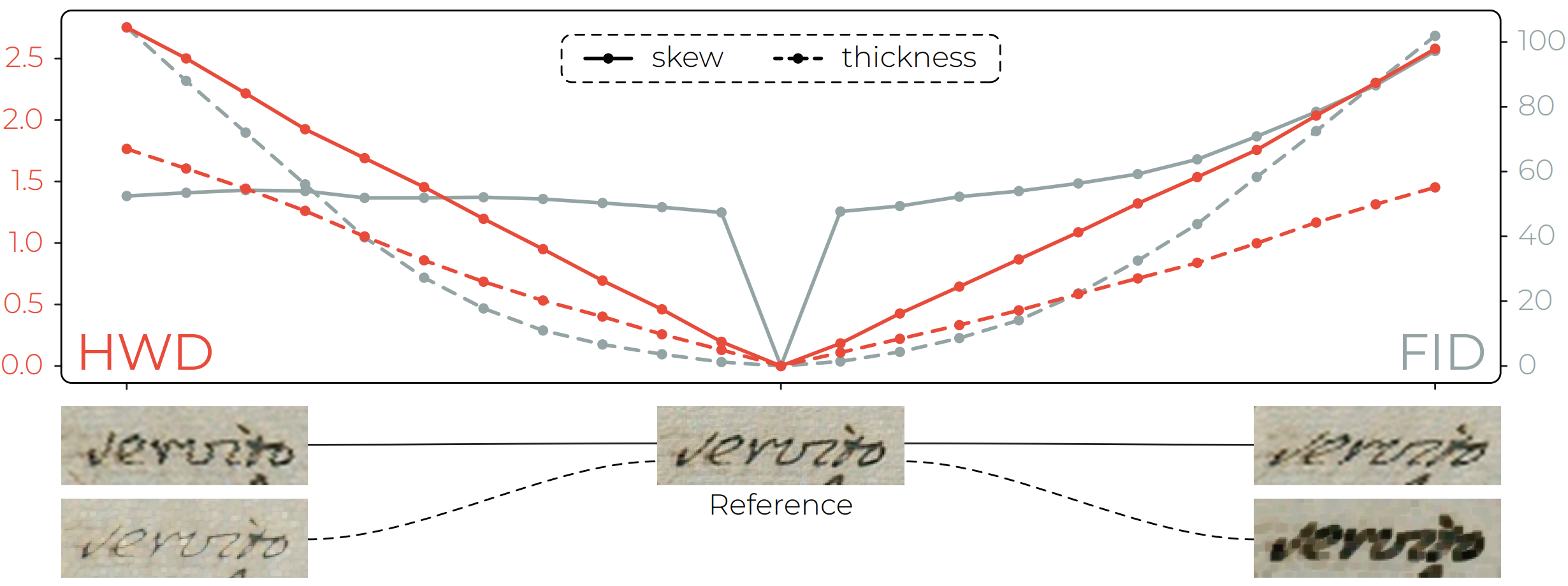}  
    \caption{FID and HWD by varying handwriting thickness and slant.}
    \label{fig:alterations}
\end{figure}

To assess the sensitivity to handwriting-related visual aspects, we compare the FID and HWD between reference images and increasingly altered ones, taken from the LAM dataset. In particular, the considered alterations entail shear, erosion, and dilation to simulate handwriting slant and strokes thickness. The results are reported in Figure~\ref{fig:alterations} and show that HWD is more sensitive than the FID to such visual aspects and that it increases linearly with the alteration intensity, thus enforcing its suitability for evaluating HTG.

\section{Conclusion}
In this work, we have proposed~\scorename, a score specifically designed to evaluate Styled HTG. \scorename~exploits the features extracted by a convolutional backbone trained on a large synthetic dataset of handwritten text to compare the perceptual differences between handwritings. Moreover, it is designed to work with images of variable lengths, such as those containing text. 
The results obtained from extensive experimental analysis demonstrate its suitability for evaluating text image generation approaches, its sensitivity to different styles, and its numerical stability. Hopefully, the use of the proposed score, whose implementation and backbone model weights will be made publicly available, will contribute to pushing forward the research on the Styled HTG task.

\clearpage 
\bibliography{main}
\end{document}

\maketitle

\section{Additional Results on the Sensitivity to Handwriting}
\begin{figure}[h]
    \centering
    \resizebox{0.65\linewidth}{!}{
    \setlength{\tabcolsep}{.1em}
    \renewcommand{\arraystretch}{.50}
    \begin{tabular}{c cc}
         & \textbf{NorHand} & \textbf{Bangla Writing} \\
         \rotatebox[origin=c]{90}{\textbf{FID}} &
         \raisebox{-0.5\height}{\includegraphics[width=0.33\textwidth]{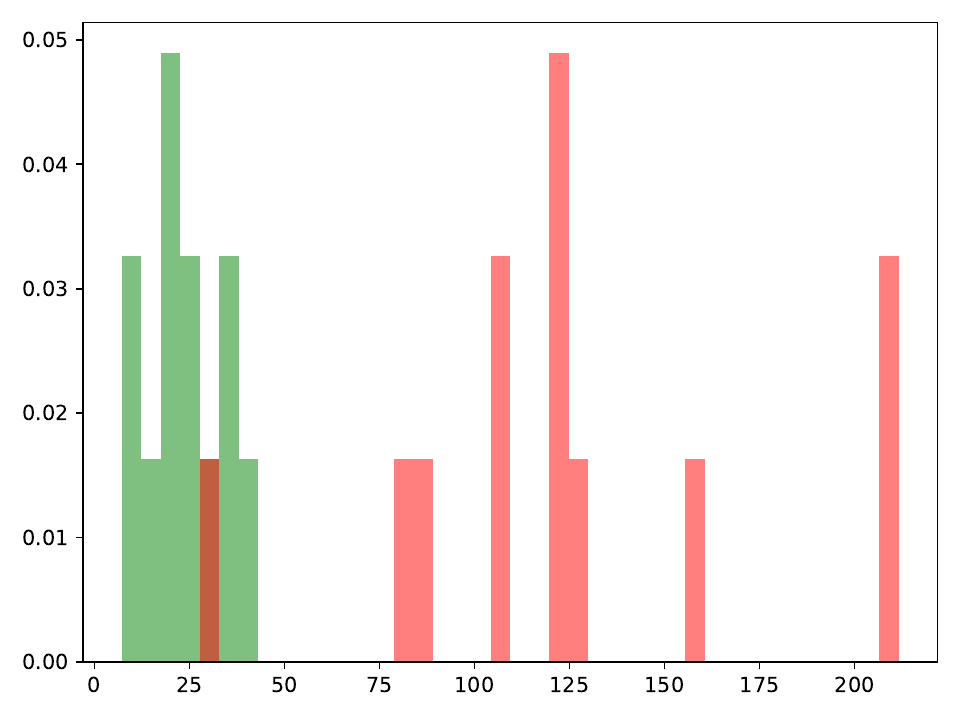}}&
         \raisebox{-0.5\height}{\includegraphics[width=0.33\textwidth]{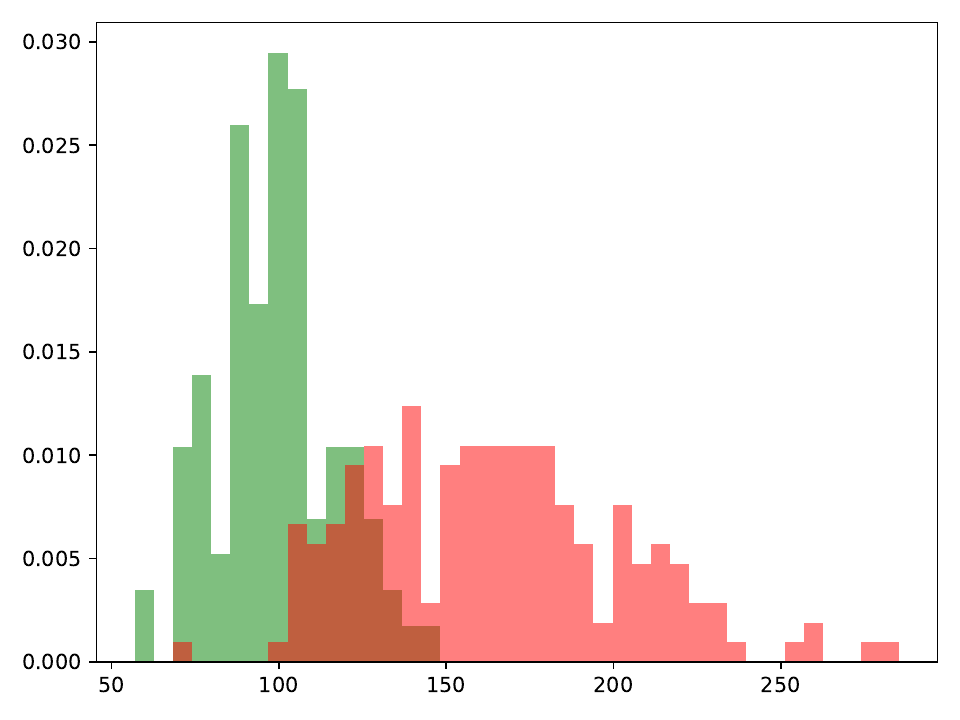}}\\ 
         \rotatebox[origin=c]{90}{\textbf{\scorename}} &
         \raisebox{-0.5\height}{\includegraphics[width=0.33\textwidth]{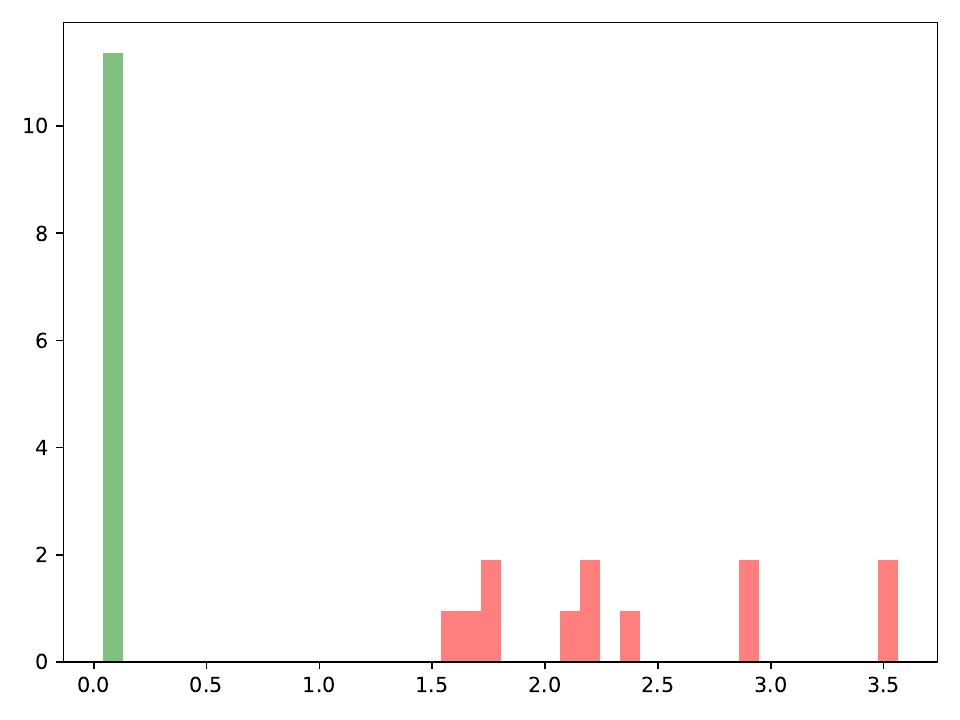}}&
         \raisebox{-0.5\height}{\includegraphics[width=0.33\textwidth]{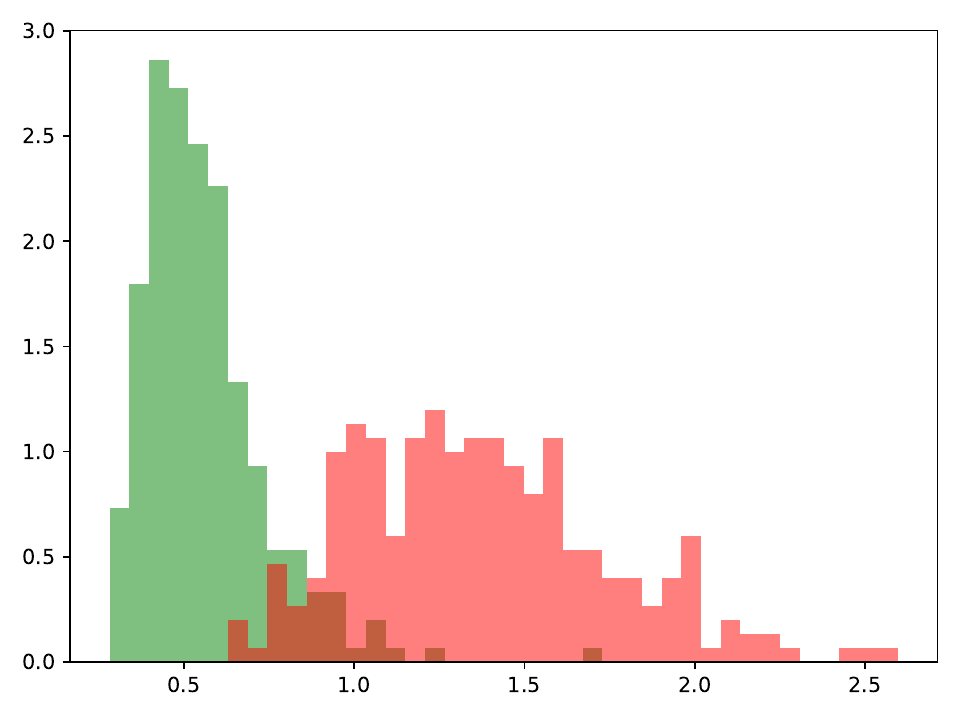}} \\
    \end{tabular}}
    \caption{Distributions of different scores used to evaluate HTG models when applied on same-author (green) or different-author (red) subsets. The overlap area is in dark red.
   }
    \label{fig:distributions_supp}
\end{figure}

In this section, we report the results on the sensitivity of the \scorename~and the FID to the handwriting style obtained on the Norhand and BanglaWriting multi-author datasets. We consider half of the samples for each featured writer as references and the other half as if they were the output of a perfect Styled HTG model. Then, we compare the distributions of the \scorename~and the FID values computed on text images of multiple matching and non-matching authors pairs.  
The obtained distributions are reported in Figure~\ref{fig:distributions_supp}. 

\section{Further Comparison Between HTG Approaches}
In Table~\ref{tab:images}, we report some qualitative examples of images generated by two HTG approaches being scored with both FID and HWD. These show that HWD better separates cases in which HTG models perform one better than the other, compared to the FID, which has similar values both for good cases and failure cases.

\begin{table}[h]
    \centering
    \setlength{\tabcolsep}{.9em}
    \resizebox{.7\linewidth}{!}{
    \begin{tabular}{c cc c cc}
            \midrule
            \multirow{2}{*}{\textbf{Reference}} & 
            \multicolumn{2}{c}{\textbf{VATr}} && 
            \multicolumn{2}{c}{\textbf{HWT}} \\
         \cmidrule{2-3} \cmidrule{5-6}
         & HWD & FID && HWD & FID \\
        \midrule
        \multirow{2}{*}{
        \includegraphics[height=0.25cm]{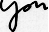} 
        \hspace{.01em}
        \includegraphics[height=0.25cm]{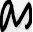}
        \hspace{.01em}
        \includegraphics[height=0.25cm]{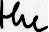}}      
        &
        \multicolumn{2}{c}{
        \includegraphics[height=0.25cm]{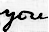} 
        \hspace{.01em}
        \includegraphics[height=0.25cm]{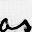}
        \hspace{.01em}
        \includegraphics[height=0.25cm]{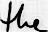}}       
        &&
        \multicolumn{2}{c}{
        \includegraphics[height=0.25cm]{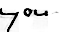} 
        \hspace{.01em}
        \includegraphics[height=0.25cm]{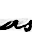}
        \hspace{.01em}
        \includegraphics[height=0.25cm]{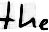}} \\
        \cmidrule{2-3} \cmidrule{5-6}
        & 0.84 & 128.4 && 1.09 & 128.7 \\
        \midrule
        \multirow{2}{*}{
        \includegraphics[height=0.25cm]{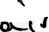} 
        \hspace{.01em}
        \includegraphics[height=0.25cm]{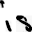}
        \hspace{.01em}
        \includegraphics[height=0.25cm]{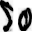}} 
        &
        \multicolumn{2}{c}{
        \includegraphics[height=0.25cm]{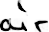} 
        \hspace{.01em}
        \includegraphics[height=0.25cm]{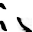}
        \hspace{.01em}
        \includegraphics[height=0.25cm]{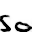}} 
        &&
        \multicolumn{2}{c}{
        \includegraphics[height=0.25cm]{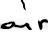} 
        \hspace{.01em}
        \includegraphics[height=0.25cm]{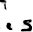}
        \hspace{.01em}
        \includegraphics[height=0.25cm]{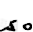}} \\
        \cmidrule{2-3} \cmidrule{5-6}
        & 0.70 & 111.7 && 0.93 & 111.9 \\
        \midrule
        \multirow{2}{*}{
        \includegraphics[height=0.25cm]{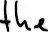}
        \hspace{.01em}
        \includegraphics[height=0.25cm]{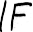}
        \hspace{.01em}
        \includegraphics[height=0.25cm]{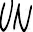}} 
        &
        \multicolumn{2}{c}{
        \includegraphics[height=0.25cm]{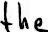} 
        \hspace{.01em}
        \includegraphics[height=0.25cm]{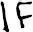}
        \hspace{.01em}
        \includegraphics[height=0.25cm]{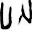}} 
        &&
        \multicolumn{2}{c}{
        \includegraphics[height=0.25cm]{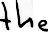} 
        \hspace{.01em}
        \includegraphics[height=0.25cm]{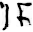}
        \hspace{.01em}
        \includegraphics[height=0.25cm]{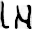}} \\
        \cmidrule{2-3} \cmidrule{5-6}
        & 0.60 & 86.6 && 0.92 & 86.7 \\
        \midrule
    \end{tabular}}
    \caption{Qualitatives, FID, and HWD of HTG models.}\label{tab:images}
\end{table}

\section{Alternative Metric Distances for HWD}
In Table~\ref{tab:distances} we report the results obtained on the IAM dataset when using Mahalanobis and the Hamming distances in the final step of the HDW computation. It emerges that the Euclidean distance works best for HWD, leading to the smaller Overlap and EER.

\begin{table}[h!]
    \centering
    \resizebox{0.4\linewidth}{!}{
    \begin{tabular}{ccc}
        \toprule
        \textbf{Distance} & \textbf{Overlap} & \textbf{EER} \\ 
        \midrule
        Mahalanobis & 7.4          & 3.6          \\
        Hamming     & 4.3          & 2.1          \\
        Euclidean   & \textbf{0.7} & \textbf{0.3} \\
        \bottomrule
    \end{tabular}}
    \caption{Ablation results when changing the distance metric at the final step of HWD.}\label{tab:distances}
\end{table}

\section{Additional Results on the Sensitivity to the Number of Samples}
\begin{figure}[h]
    \centering
    \setlength{\tabcolsep}{.1em}
    \renewcommand{\arraystretch}{.50}
    \begin{tabular}{c ccc}
         & \textbf{LAM} & \textbf{Rodrigo} & \textbf{ICFHR14} \\
         \rotatebox[origin=c]{90}{\textbf{FID}} &
         \raisebox{-0.5\height}{\includegraphics[width=0.30\textwidth]{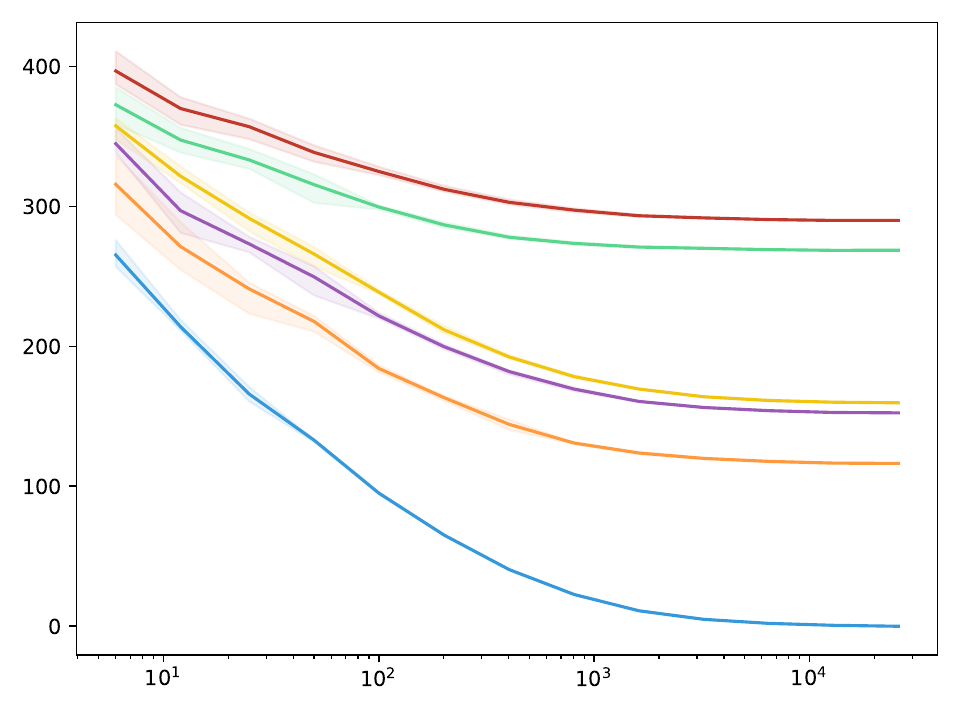}}&
         \raisebox{-0.5\height}{\includegraphics[width=0.30\textwidth]{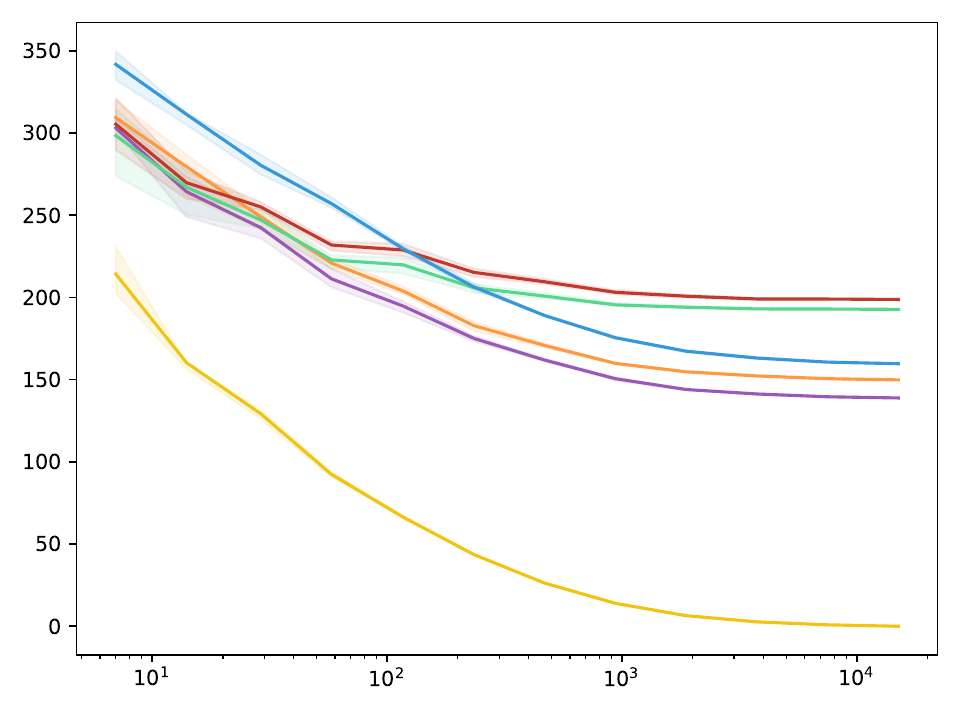}}&
         \raisebox{-0.5\height}{\includegraphics[width=0.30\textwidth]{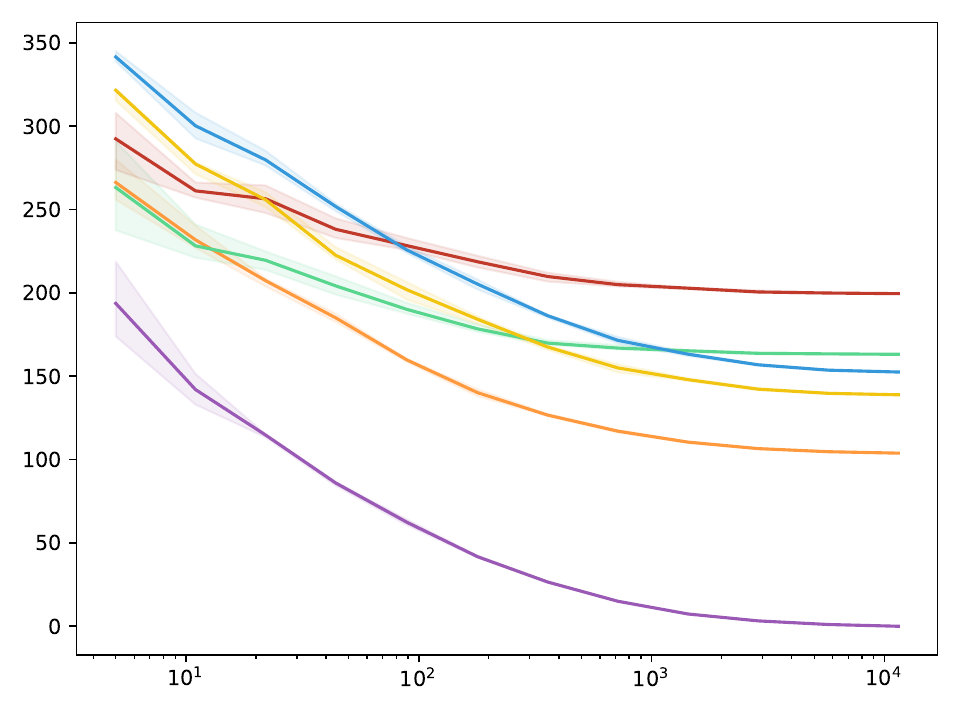}}\\
         \rotatebox[origin=c]{90}{\textbf{HWD}} &
         \raisebox{-0.5\height}{\includegraphics[width=0.30\textwidth]{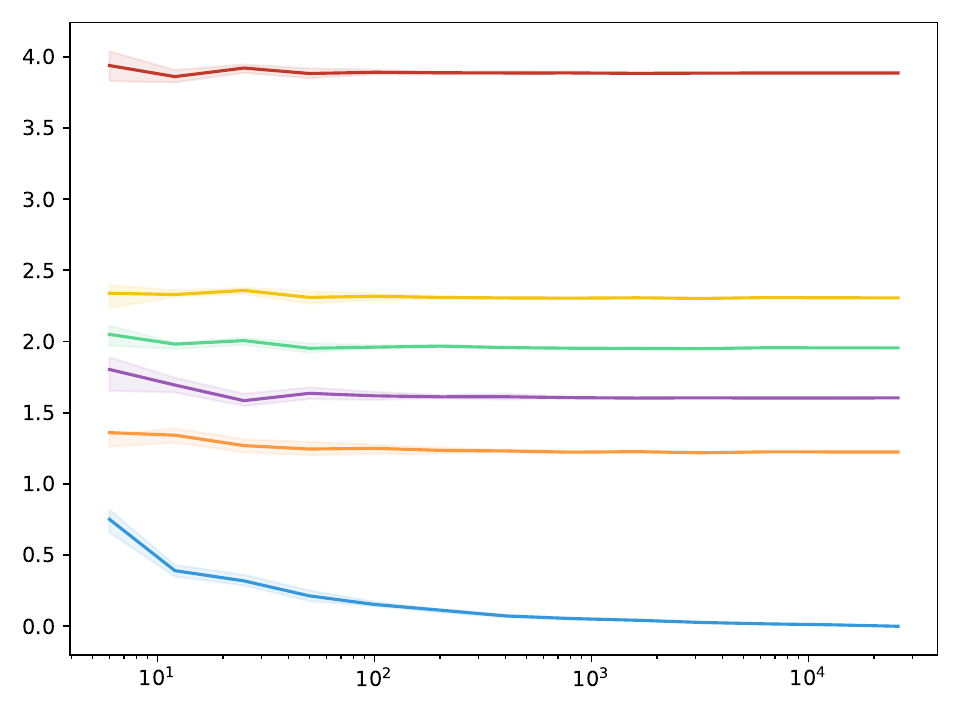}}&
         \raisebox{-0.5\height}{\includegraphics[width=0.30\textwidth]{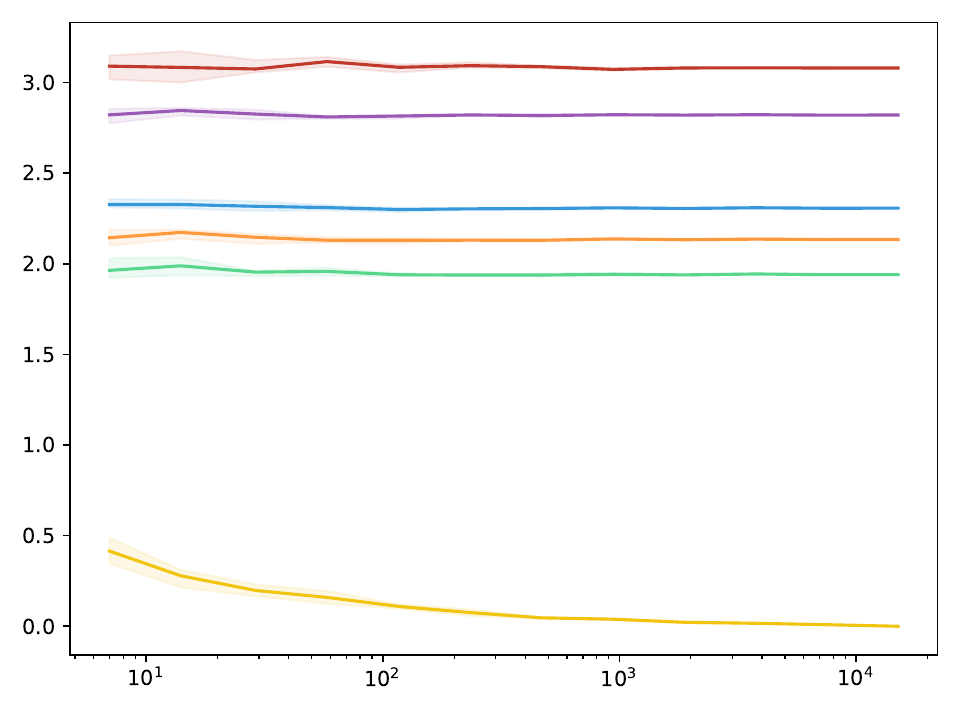}}&
         \raisebox{-0.5\height}{\includegraphics[width=0.30\textwidth]{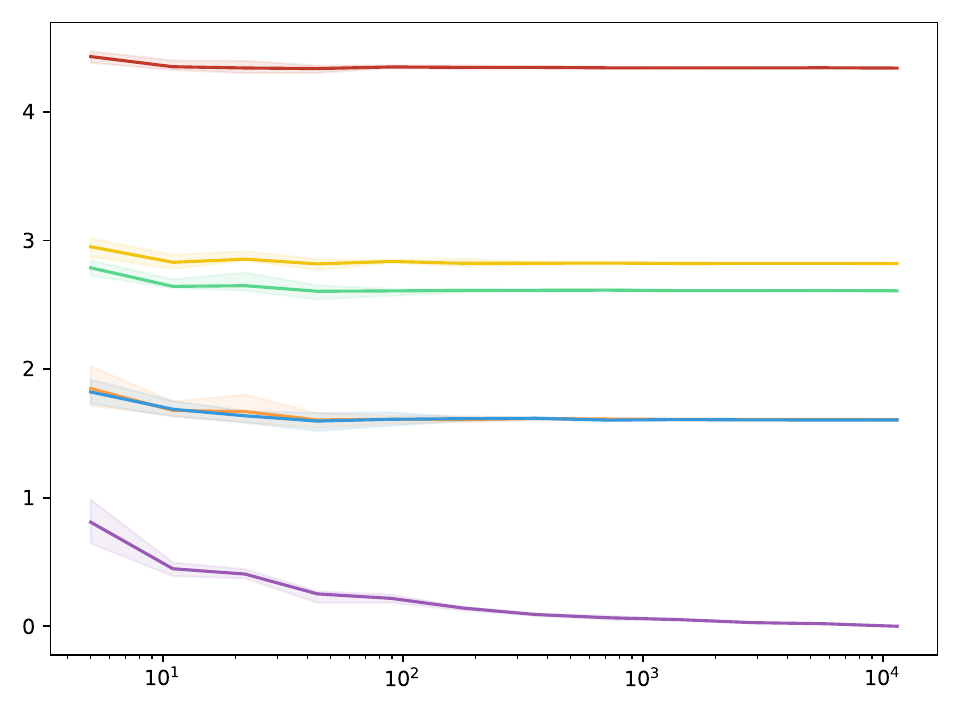}}\\ 
         \\
         & \textbf{Leopardi} & \textbf{Saint Gall} & \textbf{Washington} \\
         \rotatebox[origin=c]{90}{\textbf{FID}} &
         \raisebox{-0.5\height}{\includegraphics[width=0.30\textwidth]{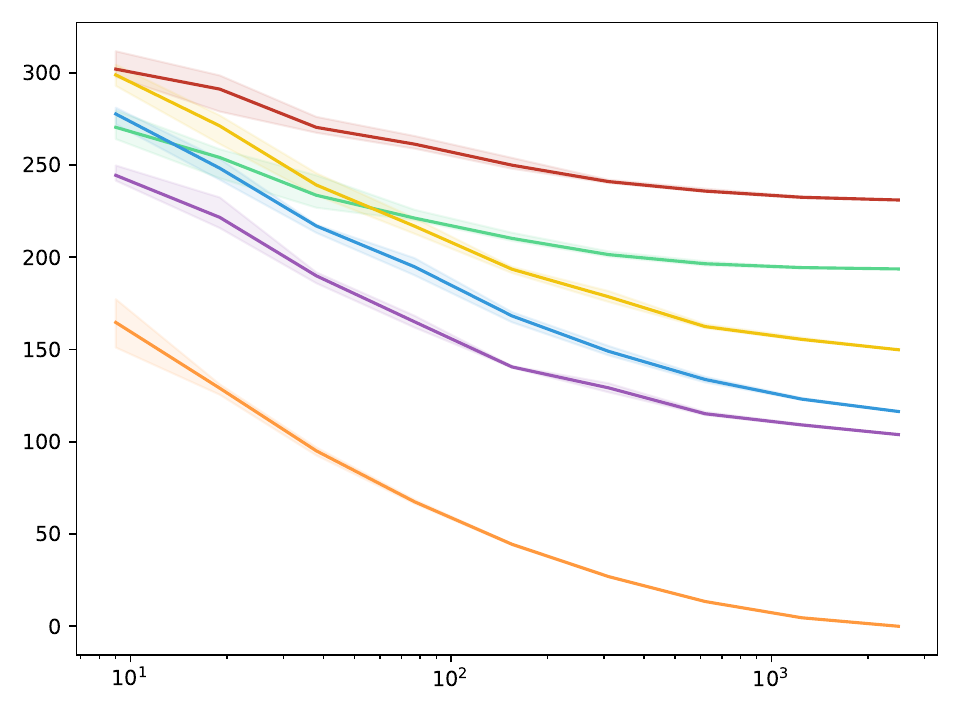}}&
         \raisebox{-0.5\height}{\includegraphics[width=0.30\textwidth]{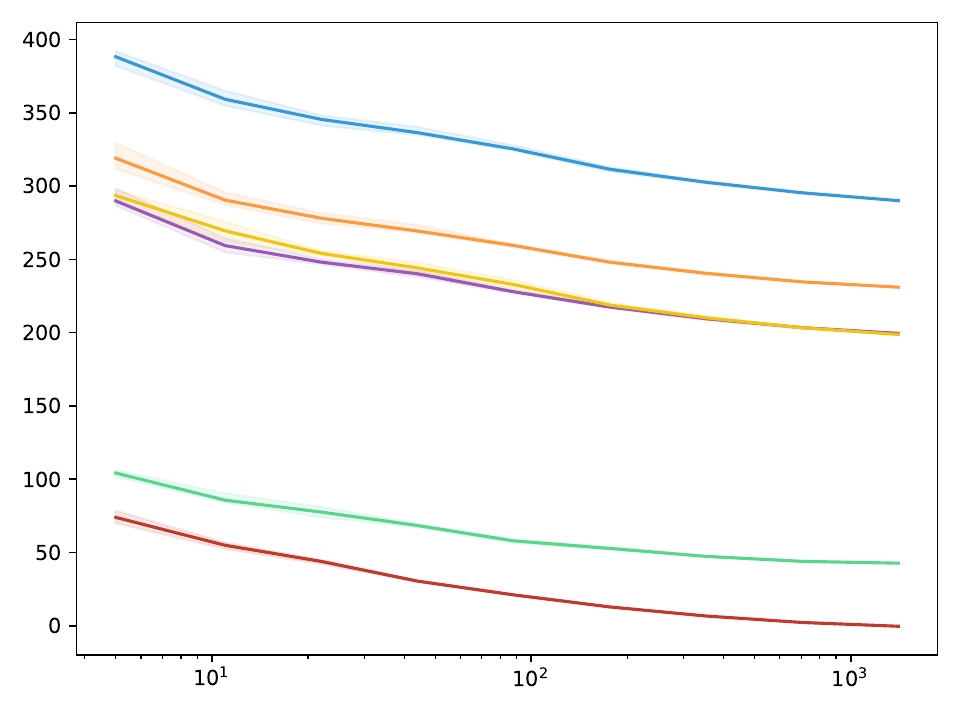}}&
         \raisebox{-0.5\height}{\includegraphics[width=0.30\textwidth]{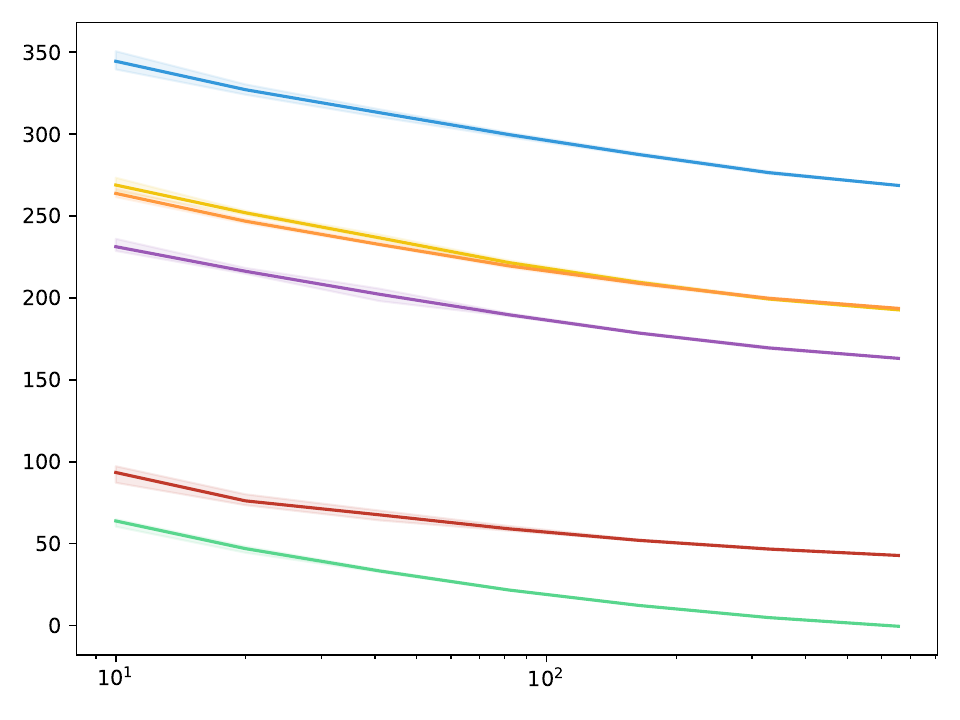}}\\
         \rotatebox[origin=c]{90}{\textbf{HWD}} &
         \raisebox{-0.5\height}{\includegraphics[width=0.30\textwidth]{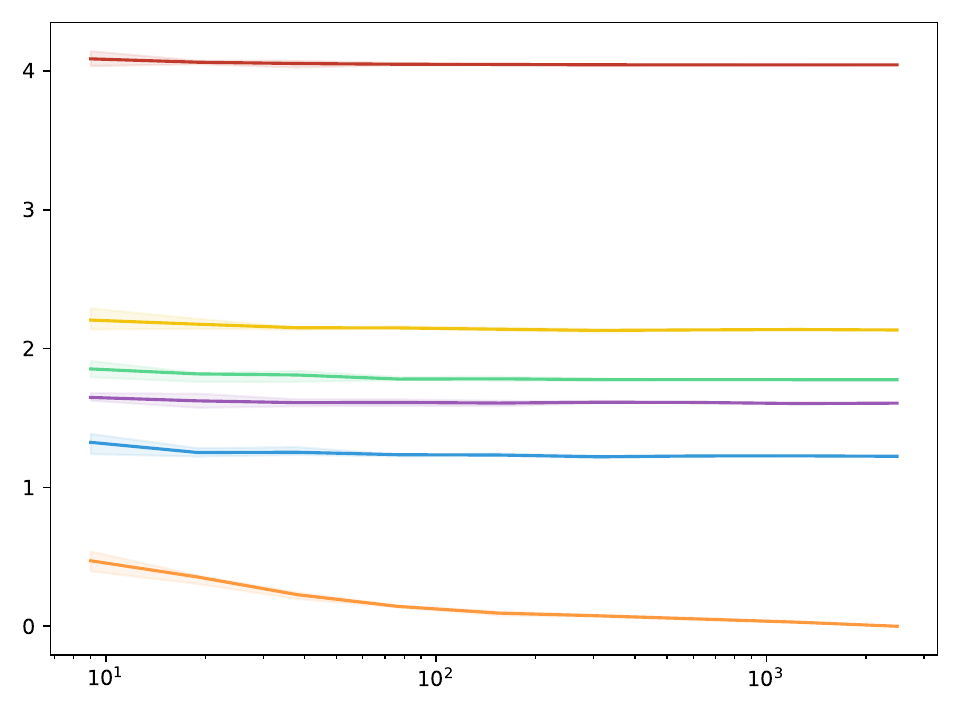}}&
         \raisebox{-0.5\height}{\includegraphics[width=0.30\textwidth]{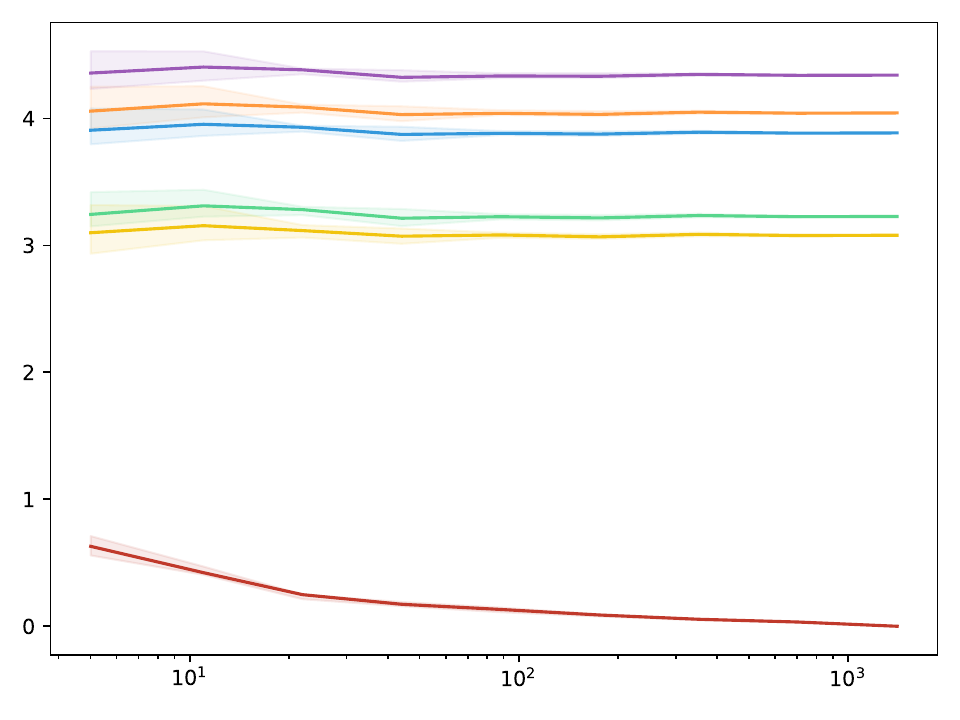}}&
         \raisebox{-0.5\height}{\includegraphics[width=0.30\textwidth]{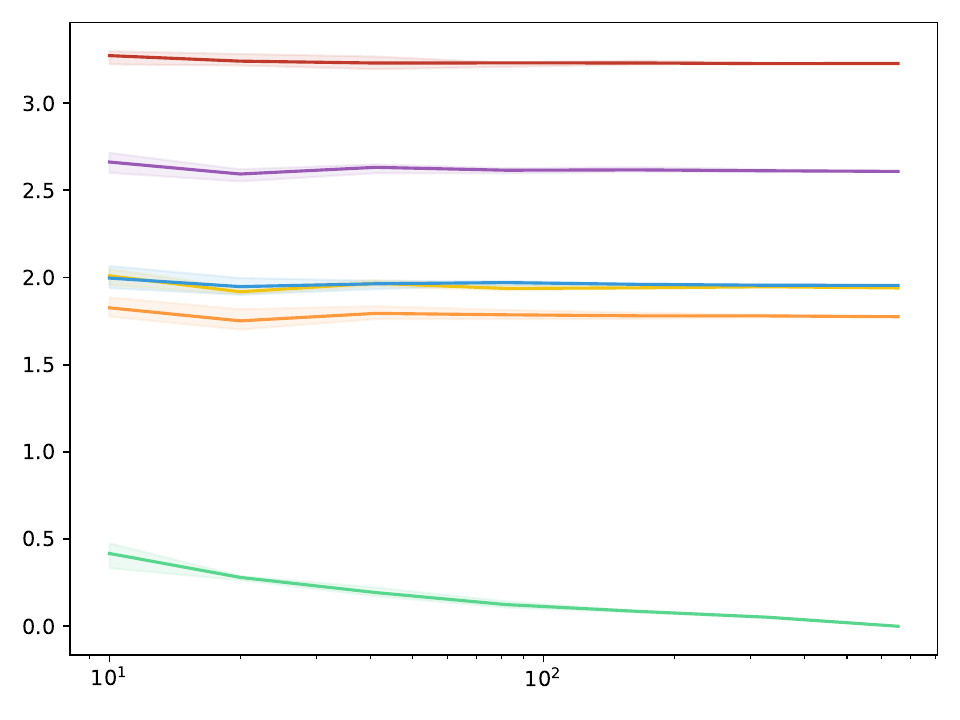}}\\
    \end{tabular}
    \\ 
    \renewcommand{\arraystretch}{1.2}
    \setlength{\tabcolsep}{.23em}
    \begin{tabular}{|cl c cl c cl c cl c cl c cl|}
    \hline
         {\includegraphics[height=0.25cm]{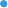}}& LAM &&
         {\includegraphics[height=0.25cm]{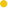}}& Rodrigo &&
         {\includegraphics[height=0.25cm]{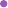}}& ICFHR14 &&
         {\includegraphics[height=0.25cm]{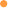}}& Leopardi &&
         {\includegraphics[height=0.25cm]{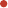}}& Saint Gall &&
         {\includegraphics[height=0.25cm]{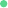}}& Washington \\       
    \hline
    \end{tabular}
    \caption{Comparison between FID and \scorename~with varying number of samples on different single-author datasets. The lines denote the mean, and the transparent bands represent the range between the 25th and 75th percentiles, obtained with 10 calculation runs.
   }
    \label{fig:sample_size_supp}
\end{figure}

In this section, we report further results on the numerical stability of the proposed \textbf{\scorename}, compared to the \textbf{FID} and two baseline scores, namely the \textbf{FID w/ Euclidean} (obtained by computing the Euclidean distance on the Inception-v3 features) and \textbf{HWD w/ Fréchet} (obtained by computing the Fréchet distance on the VGG16). In particular, we use the images from the considered single-author datasets (ICFHR14, Saint Gall, Leopardi, Rodrigo, Washington, and LAM). The results are expressed as mean and range between the 25th and 75th percentiles of the values obtained over multiple runs by varying the number of samples. These are reported in Figure~\ref{fig:sample_size_supp}. For the plots in each row, we consider the whole indicated dataset as the set of reference images and compute the score when comparing it with a variable number of samples from the other datasets and from the dataset itself for reference.

\section{Computation Time Comparison}
We consider the computation time, consisting of image representation and distance computation, of FID and HDW on the same hardware and data. Computing the FID score on 25823/25823 real/fake images from the LAM dataset takes 426.12s + 9.03s (image representation + distance computation), while the computation time of HWD is 135.50s + 0.01s.

% \clearpage 
% \bibliography{suppl}